\documentclass[sigconf]{acmart}
\usepackage{multirow}
\usepackage{colortbl}
\usepackage{xcolor}
\usepackage{ragged2e}
\AtBeginDocument{%
  }

\copyrightyear{2025}
\acmYear{2025}
\setcopyright{cc}
\setcctype{by}
\acmConference[MM '25]{Proceedings of the 33rd ACM International Conference on Multimedia}{October 27--31, 2025}{Dublin, Ireland}
\acmBooktitle{Proceedings of the 33rd ACM International Conference on Multimedia (MM '25), October 27--31, 2025, Dublin, Ireland}\acmDOI{10.1145/3746027.3755074}
\acmISBN{979-8-4007-2035-2/2025/10}

\begin{document}

\title[B4DL]{B4DL: A Benchmark for 4D LiDAR LLM in Spatio-Temporal Understanding}

\author{Changho Choi}
\authornote{Both authors contributed equally to this research.}
\affiliation{%
  \institution{Korea Advanced Institute of Science and Technology}
  \city{Daejeon}
  \country{Republic of Korea}
}
\email{ccho4702@kaist.ac.kr}

\author{Youngwoo Shin}
\authornotemark[1]
\affiliation{%
  \institution{Korea Advanced Institute of Science and Technology}
  \city{Daejeon}
  \country{Republic of Korea}
}
\email{yshin0917@kaist.ac.kr}

\author{Gyojin Han}
\affiliation{%
  \institution{Korea Advanced Institute of Science and Technology}
  \city{Daejeon}
  \country{Republic of Korea}
}
\email{hangj0820@kaist.ac.kr}

\author{Dong-Jae Lee}
\affiliation{%
  \institution{Korea Advanced Institute of Science and Technology}
  \city{Daejeon}
  \country{Republic of Korea}
}
\email{jhtwosun@kaist.ac.kr}

\author{Junmo Kim}
\authornote{Corresponding author.}
\affiliation{%
  \institution{Korea Advanced Institute of Science and Technology}
  \city{Daejeon}
  \country{Republic of Korea}
}

\email{junmo.kim@kaist.ac.kr}

\renewcommand{\shortauthors}{Choi et al.}

\begin{abstract}
  Understanding dynamic outdoor environments requires capturing complex object interactions and their evolution over time. LiDAR-based 4D point clouds provide precise spatial geometry and rich temporal cues, making them ideal for representing real-world scenes. However, despite their potential, 4D LiDAR remains underexplored in the context of Multimodal Large Language Models (MLLMs) due to the absence of high-quality, modality-specific annotations and the lack of MLLM architectures capable of processing its high-dimensional composition. To address these challenges, we introduce \textbf{B4DL}, a new benchmark specifically designed for training and evaluating MLLMs on 4D LiDAR understanding. In addition, we propose a scalable data generation pipeline and an MLLM model that, for the first time, directly processes raw 4D LiDAR by bridging it with language understanding. Combined with our dataset and benchmark, our model offers a unified solution for spatio-temporal reasoning in dynamic outdoor environments. We provide rendered 4D LiDAR videos, generated dataset, and inference outputs on diverse scenarios at: \url{https://github.com/ccho4702/B4DL}
\end{abstract}


\begin{teaserfigure}
    \centering
  \includegraphics[width=0.9\textwidth]{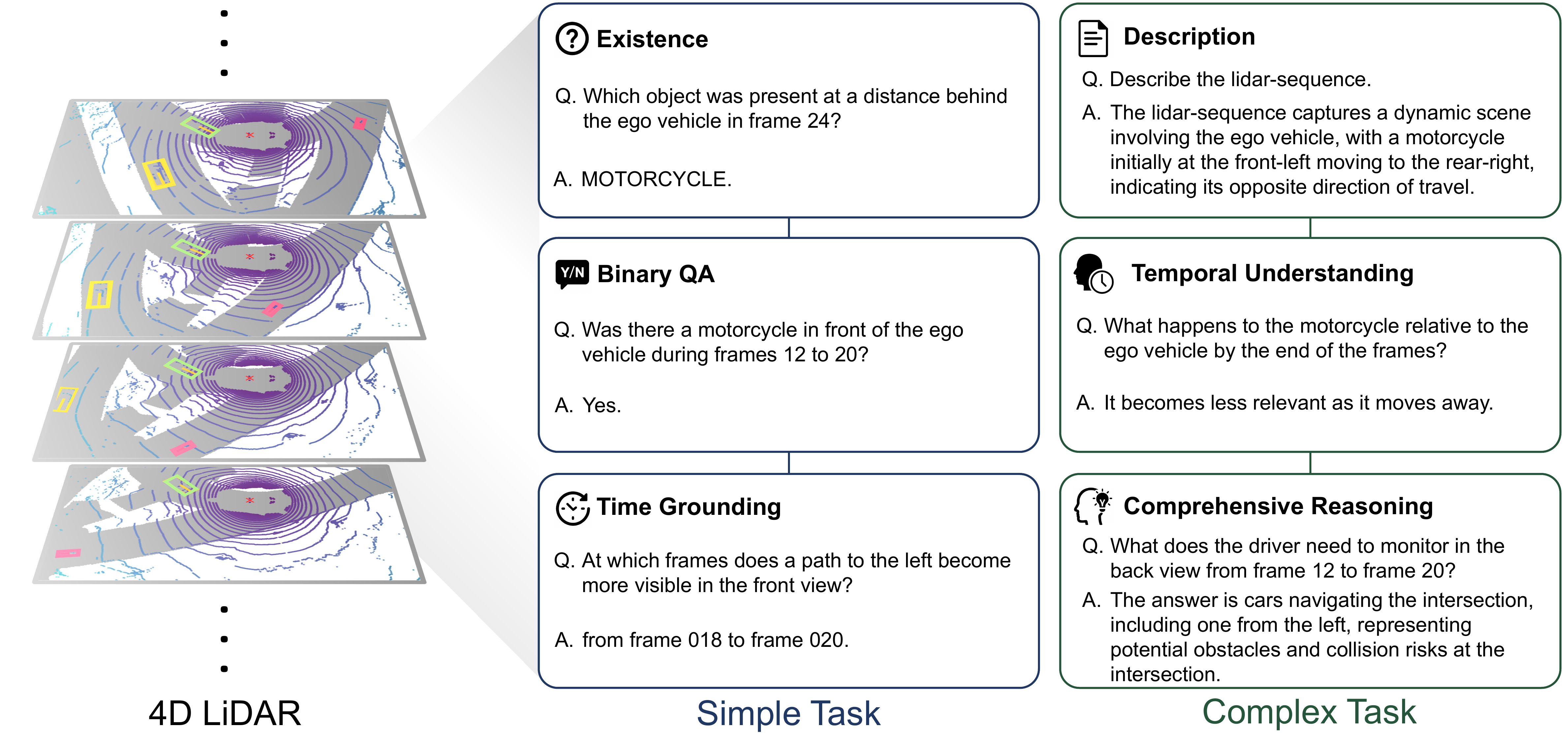}
  \caption{
  \textbf{Examples of question-answer (QA) pairs for the six B4DL tasks. The QA pairs are generated from the 12th to 24th frames of 4D LiDAR illustrated on the left using our proposed data generation pipeline. In the visualization of the 4D LiDAR, the pink and green markers indicate the locations of the motorcycle and car mentioned in the example QA pairs, respectively.}}
  \Description[]{We generate the dataset using our data generation pipeline.}
  \label{fig:overview}
\end{teaserfigure}


\maketitle

\section{Introduction}

Understanding outdoor scenes in real-world scenarios requires precisely capturing the complexities of an evolving environment, such as changing object interactions, environmental variations, and motion patterns over time. Many researchers \cite{zhang2022pointclip, zhu2023pointclipv2, xue2024ulip, hegde2023clip, hong20233d} have dedicated their efforts to this task, emphasizing the importance of accurately interpreting dynamic environments. To understand how an outdoor scene is structured, Light Detection and Ranging (LiDAR), which captures object orientations and geometries omnidirectionally, is widely used in many large-scale datasets\cite{sun2020scalability, geiger2013vision, caesar2020nuscenes}. When the point clouds captured by LiDAR are accumulated over time, they form 4D LiDAR, a modality consisting of sequential LiDAR point clouds that provides a rich temporal structure and spatial orientation to the dataset. These scenes not only reflect the spatial environments we perceive but also capture temporal changes, closely resembling the dynamic reality humans experience. 

Although the 4D LiDAR \cite{ma2024cam4docc, paek2022k} offers a highly accurate depiction of real-world scenes, and recent Multimodal Large Language Models (MLLMs) have demonstrated strong capabilities in understanding dynamic outdoor environments, integrating them remains largely underexplored. A key obstacle is the absence of publicly available 4D LiDAR datasets with modality-specific textual annotations. This is due to the difficulty of constructing such datasets. Manual annotation is prohibitively expensive at scale, while automated generation using LLMs is limited by their lack of inductive bias toward high-dimensional LiDAR data. Furthermore, even if annotated datasets were available, existing MLLM architectures are not designed to directly handle raw 4D LiDAR. These limitations pose a fundamental barrier to enabling MLLMs to reason over dynamic scenes using the rich information encoded in 4D LiDAR.

To address these challenges, we introduce \textbf{B4DL}, a new benchmark designed for 4D LiDAR LLMs to understand the spatio-temporal dynamics of 4D LiDAR in its high-dimensional form by leveraging LLMs. To facilitate the understanding of 4D dynamics in outdoor scenes, we define two sets of tasks presented in Figure~\ref{fig:overview}. The first set includes simple tasks that emphasize the model’s ability to provide short, precise responses, demonstrating high-level comprehension of outdoor scenes captured by 4D LiDAR. These tasks include Existence, Binary QA, and Time Grounding. The second set comprises more complex tasks that require the model to answer questions about spatio-temporal dynamics using more elaborate and sophisticated language, thereby evaluating its deeper understanding of 4D scenes. These tasks include Description, Temporal Understanding, and Comprehensive Reasoning. 

To support these tasks, we design a novel data generation pipeline capable of producing high-quality textual datasets tailored for 4D LiDAR. The pipeline consists of two key steps: first, extracting high-level semantic information from multi-view images aligned with LiDAR data; second, transforming this information into diverse, instruction-style formats suitable for training MLLMs. By explicitly separating the process of extracting semantic information from the process of generating textual data, our pipeline ensures that each step can be optimized independently. This decoupling allows for more precise control over data quality and flexibility in utilizing multimodal sources. Notably, the pipeline is applicable to any dataset that contains synchronized LiDAR-camera pairs, enabling broad generalization across diverse sources of spatio-temporal LiDAR data. Using this pipeline, we construct a synthetic textual dataset, the \textbf{B4DL dataset}, based on LiDAR point clouds from the nuScenes dataset \cite{caesar2020nuscenes}. This is the first publicly available textual dataset for 4D LiDAR, containing 178.4k samples, each consisting of paired question-and-answer (QA) data designed to enable MLLMs to understand 4D scenes in both spatial and temporal contexts.

Finally, we propose the \textbf{B4DL model}, an MLLM designed to interpret real-world spatio-temporal environments by leveraging 4D LiDAR. The B4DL model incorporates three alignment modules: \textit{Encoder for LiDAR Point Cloud, LiDAR-Text Aligner}, and \textit{Metatoken}. These modules enable the MLLM to process high-dimensional 4D LiDAR data by capturing spatial structures from point clouds and aligning them with the embedding space of the language model. To train the model effectively, we adopt a stage-wise learning strategy consisting of two stages: \textit{3D LiDAR Understanding Stage} which focuses on learning spatial features from static point clouds, and \textit{4D LiDAR Understanding Stage} which extends the model's capability to capture temporal dynamics and reason over 4D LiDAR data. The detailed contributions of our work are as follows:

\begin{itemize}
\item We introduce \textbf{B4DL}, a benchmark designed for MLLMs to understand spatio-temporal dynamics in 4D LiDAR data, with a suite of tasks that evaluate both simple and complex understanding of outdoor scenes.
\item We design a new data generation pipeline that produces instruction-style datasets from sequentially captured LiDAR data, and use it to construct the \textbf{B4DL dataset}, the first publicly available large-scale synthetic dataset with 178.4k question-answer pairs based on nuScenes.
\item We propose the \textbf{B4DL model}, an MLLM for processing high-dimensional 4D LiDAR data, along with a stage-wise training pipeline that enhances spatio-temporal reasoning.
\end{itemize}

\begin{table*}[ht]
    \centering
    \caption{Comparison of existing LiDAR datasets. 
    designed for outdoor scene understanding, emphasizing key characteristics such as omnidirectional coverage, temporal sequence availability, and suitability for MLLM training. Note that Multi-Frame refers to the ability to process more than one frame but does not necessarily imply a sequential nature. }
    \renewcommand{\arraystretch}{1.3}  
    \renewcommand{\tabcolsep}{5pt}  
    \resizebox{0.9\textwidth}{!}{  
    \begin{tabular}{|l|c|c|c|c|c|c|c|c}
        \toprule
        \textbf{Dataset} & \textbf{Input} & \textbf{LiDAR-Specific} & \textbf{360° Coverage} & \textbf{Multi-Frame} & \textbf{Sequence} & \textbf{Annotation for Sequence} & \textbf{Suitable for MLLM Training} \\
        \midrule
        \textbf{DriveLM \cite{sima2023drivelm}} & Camera & X & O & O & X & X & O \\
        \textbf{LingoQA \cite{marcu2024lingoqa}} & Camera & X & X & O & O & O & O \\
        \textbf{DriveGPT4 \cite{xu2024drivegpt4}} & Camera & X & X & O & O & O & O \\
        \textbf{nuScenes-QA \cite{qian2023nuscenes}} & Camera+LiDAR & O & O & O & O & X & X \\
         \textbf{LiDARLLM \cite{yang2023lidar}} & LiDAR & O & O & X & X & X & O \\
        \midrule
        \textbf{B4DL(Ours)} & LiDAR & O & O & O & O & O & O \\
        \bottomrule
    \end{tabular}
    }
     
    \Description[]{We made an effort to include a wide variety of LiDAR datasets in our comparison.}
    \label{tab:dataset_comparison}
\end{table*}

\section{Related Works}

\subsection{LiDAR dataset}

LiDAR datasets have played a central role in advancing 3D perception tasks in autonomous driving and robotics. KITTI \cite{geiger2013vision}, ScanNet \cite{dai2017scannet}, and Waymo \cite{sun2020scalability} have provided large-scale 3D point cloud data for tasks such as depth estimation, object detection, and semantic segmentation. To support higher-level reasoning, datasets like nuScenes \cite{caesar2020nuscenes} and ONCE \cite{mao2021one} have introduced extensive annotations and synchronized multimodal data. Notably, nuScenes collects 3D LiDAR with six RGB cameras across sequences, enabling multimodal understanding in urban environments. Recent efforts such as LiDAR-LLM \cite{yang2023lidar}, nuScenes-QA \cite{qian2023nuscenes}, and DriveLM \cite{sima2023drivelm} have begun leveraging large language models (LLMs) by pairing 3D LiDAR point clouds with textual descriptions or QA pairs to support language-based 3D reasoning. However, these datasets and methods focus predominantly on static spatial understanding from single frames. In contrast, our work introduces a dataset that incorporates temporal sequences of LiDAR, extending the understanding from 3D to 4D spatio-temporal dynamics. By modeling time as an explicit dimension, we enable reasoning over dynamic changes in real-world environments. Table \ref{tab:dataset_comparison} summarizes key differences between our dataset and existing LiDAR-based datasets.

\subsection{Multimodal Large Language Models}

Large Language Models (LLMs) such as GPT \cite{achiam2023gpt}, LLaMA \cite{grattafiori2024llama}, and Gemini \cite{team2023gemini} have achieved impressive performance in natural language understanding by modeling complex context across sentences and documents. To extend their reasoning capabilities beyond text, recent research has focused on multimodal, which incorporate additional input modalities. A central challenge in MLLMs is modality alignment, which involves mapping non-textual inputs into a representation space interpretable by LLMs. Contrastive learning has been a widely adopted approach for this, with CLIP \cite{radford2021learning} aligning images and text, and follow-up works like PointCLIP \cite{zhang2022pointclip, zhu2023pointclipv2} and LiDARCLIP \cite{hess2024lidarclip} extending this paradigm to 3D point clouds and LiDAR data. Other modalities such as audio and video have also been integrated into LLMs, as demonstrated by AudioGPT \cite{huang2024audiogpt} and video-centric models like Video-ChatGPT \cite{maaz2023videochatgpt} and VTimeLLM \cite{huang2024vtimellm}. These approaches effectively model temporal sequences, but lack 3D spatial understanding of real-world environments. Conversely, recent 3D-LLMs \cite{hong20233d} enable spatial reasoning over static point clouds, but typically do not incorporate temporal sequences. As a result, existing MLLMs are limited in their ability to jointly model both space and time, which is an essential aspect of dynamic real-world perception.

\section{Benchmark and Dataset}
\label{sec:dataset}
To rigorously evaluate scene understanding capabilities in the context of 4D LiDAR data, we design a benchmark encompassing a diverse set of tasks. The proposed benchmark is designed to reflect the core challenges of LiDAR-based spatio-temporal reasoning by accounting for issues that may arise in real-world driving scenarios such as complex object dynamics and interactions. The benchmark includes tasks spanning both simple and complex reasoning levels, such as binary object presence checks, relational question answering, temporal event understanding, and free-form scene description.

To construct a large-scale dataset for these tasks, we utilize nuScenes \cite{caesar2020nuscenes} as the data source. Our data generation pipeline leverages multi-view camera images as intermediaries to extract language-expressive 4D LiDAR context using GPT-4o \cite{achiam2023gpt}, and complements this with structured human annotations as the ground truth. This pipeline ensures both rich linguistic expressiveness and accurate spatio-temporal grounding. We further present task-specific evaluation metrics and detailed dataset statistics to highlight the scale and diversity of the benchmark.

\subsection{Task Definition}
\label{ssec:task}
To evaluate the scene understanding capability of models on 4D LiDAR data, we design a benchmark comprising two task groups: \textit{Simple Tasks} and \textit{Complex Tasks}, each targeting different levels of reasoning over spatio-temporal information.

\noindent\textbf{Simple Tasks} include Existence, Binary QA, and Time Grounding. Existence evaluates whether a model can detect the presence or absence of a specific object or class within a scene. Binary QA covers factual questions with binary (yes/no) answers, which require perceptual grounding and basic relational reasoning. Time Grounding assesses the model’s ability to localize a specific event or activity within a temporal window given a corresponding query.

\noindent\textbf{Complex Tasks} consist of Description, Temporal Understanding, and Comprehensive Reasoning. Description evaluates whether a model can generate a coherent and informative summary of a scene, capturing key entities and their dynamics. Temporal Understanding focuses on a model’s capability to reason about events and interactions that evolve over time, requiring temporal association and dynamic relational reasoning. Lastly, Comprehensive Reasoning evaluates a model’s capacity to address diverse and open-ended questions by interpreting the overall context of the scene.

\noindent\textbf{Evaluation metrics} are set to align with the objectives of each task type and to fairly assess the model’s performance. For the \textit{Simple Tasks}, we use accuracy for Existence and Binary QA, evaluated based on exact answer matching, and mIoU for Time Grounding to measure overlap between predicted and ground-truth temporal segments. For the \textit{Complex Tasks}, including Description, Temporal Understanding, and Comprehensive Reasoning, we employ BLEU-4 (B@4) \cite{papineni2002bleu}, METEOR \cite{banerjee2005meteor}, ROUGE-L \cite{lin2004rouge}, and BERTScore \cite{devlin2019bert} to evaluate both linguistic quality and semantic relevance. We also use GPT-4o \cite{achiam2023gpt} as a reference-free evaluator to score coherence, relevance, and correctness on a 100-point scale.

\subsection{Data Generation Pipeline}
We construct the dataset using a two-stage pipeline that generates question–answer (QA) pairs for each sequence, a temporally segmented portion of a scene. In the first step, the \textit{4D LiDAR Context Extraction Step}, we extract natural language descriptions that reflect the spatio-temporal context of each sequence. In the second step, the \textit{Context-to-QA Transformation Step}, the extracted context is transformed into meaningful QA pairs suitable for MLLMs.

\begin{figure}[t]
    \centering
    \includegraphics[width=0.35\textwidth]{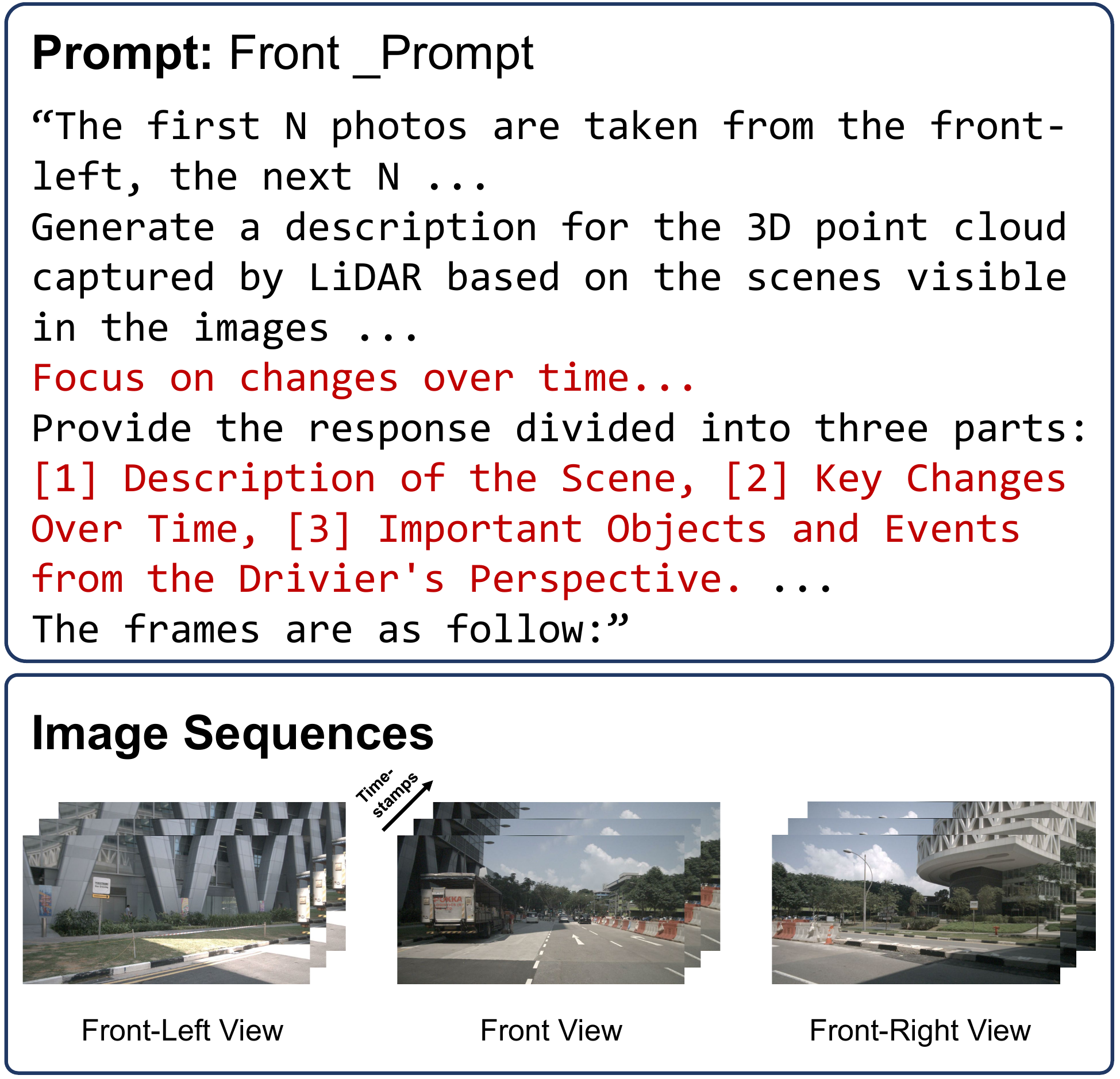}
    \caption{An example prompt guiding GPT to describe 4D LiDAR scenes using aligned frontal multi-view images, structured into scene overview, temporal changes, and ego-vehicle-centric events for spatio-temporal context extraction.
    \Description[]{Fully described in the text.}
}
    \label{fig:prompt}
\end{figure}

\noindent\textbf{4D LiDAR Context Extraction Step} generates natural language descriptions that capture the spatio-temporal dynamics of 4D outdoor scenes using GPT-4o \cite{achiam2023gpt}. However, since GPT-4o cannot directly process LiDAR input, we use six synchronized multi-view images (front, front-right, front-left, back, back-right, back-left) per frame from nuScenes \cite{caesar2020nuscenes}. Each image is temporally aligned with the corresponding LiDAR frame captured at the same timestamp. These images serve as a visual proxy for the 4D LiDAR input, enabling GPT-4o to generate coherent descriptions of the surrounding environment, object interactions, and spatial relations. Due to the limited capacity of GPT to process multiple image sequences simultaneously, the views are split into frontal and rear groups, resulting in two descriptions per sequence. The prompt used in this step explicitly guides GPT to extract LiDAR-relevant information such as object positions, orientations, distances, and movements. This approach enables the extraction of LiDAR-specific information by integrating contextual cues from surrounding images. Detailed prompt used in this step are shown in Figure \ref{fig:prompt}.

However, fully relying on even the most advanced LLMs, such as GPT, remains problematic. Despite their impressive generative capabilities, issues like factual inaccuracies, hallucinations, and ambiguity still persist. To address these limitations and improve the reliability of the generated descriptions, we additionally incorporate human annotations from nuScenes. These annotations typically include numeric details such as event timestamps, frame identifiers, object categories, and status labels. To integrate this numeric data into natural language narratives, we manually translate these annotations into descriptive phrases. For example, spatial coordinate changes are articulated as "lateral movements to the right," and object motions relative to the ego vehicle are described using terms like "moving away from the ego vehicle". This manual conversion process ensures that the resulting textual descriptions accurately and intuitively represent dynamic scene information.

\begin{figure}[t]
    \centering
    \includegraphics[trim=0cm 0cm 0cm 0cm, clip, width=0.47\textwidth]{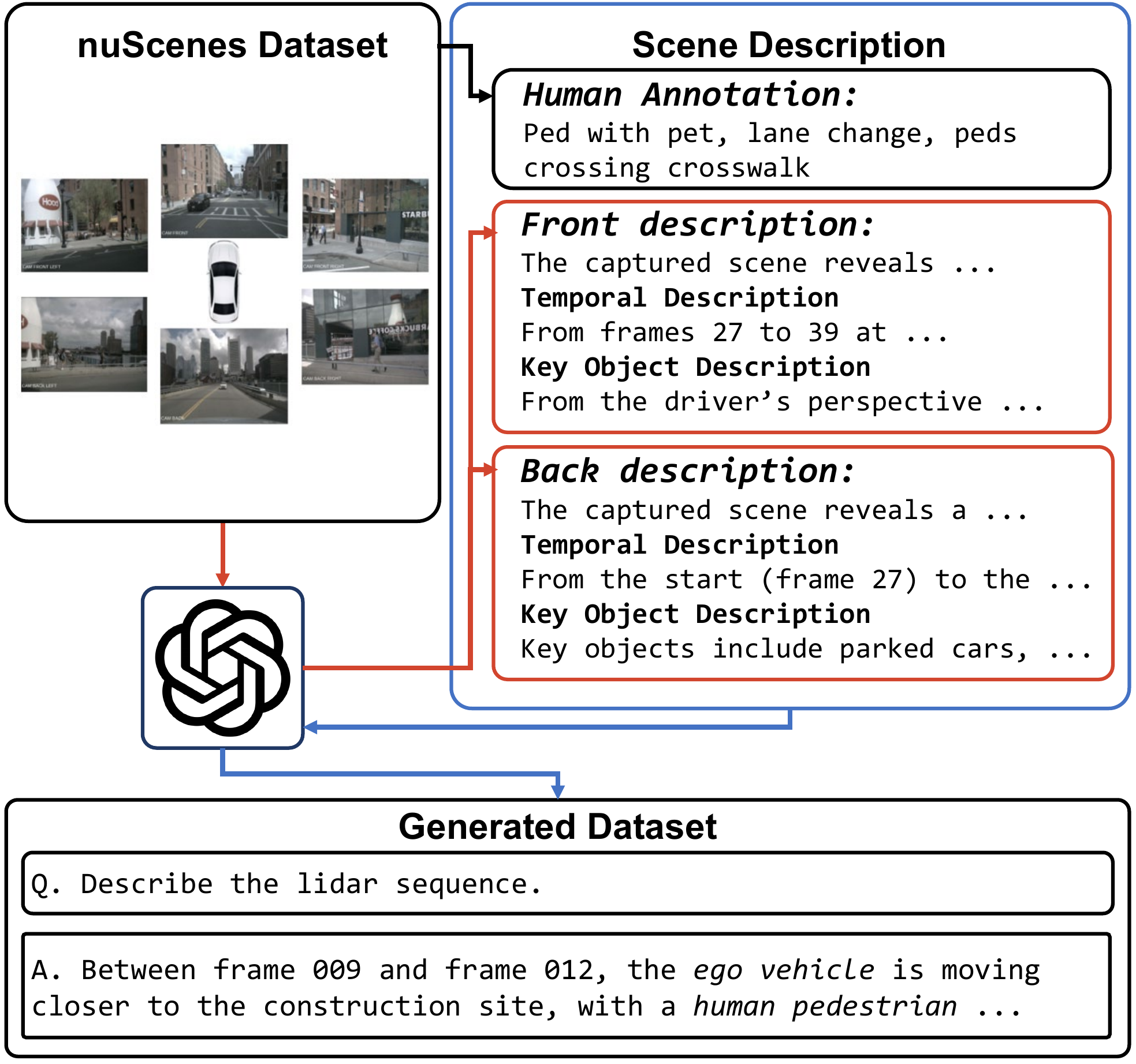}
    \caption{Overview of the dataset generation pipeline
    The six multi-view images, aligned with LiDAR point clouds, are fed into GPT with a refined prompt to generate rich scene descriptions (red line), and then the generated descriptions are processed again to create QA pairs, ensuring a LiDAR-centric perspective (blue line).}
    \Description[]{he generation process was designed with consideration for the input sequence length limitations and question types that GPT models can effectively handle.}
    \label{fig:pipeline}
\end{figure}

\noindent\textbf{Context-to-QA Transformation Step} focuses on constructing diverse and meaningful QA pairs that encourage models to understand the spatio-temporal dynamics of 4D LiDAR sequences. Instead of relying on fixed templates or handcrafted rules, we leverage GPT to generate context-aware QA pairs by jointly using the synthetic textual descriptions from front and rear views obtained in the previous step, along with structured human annotations. To ensure relevance and diversity, we design task-specific prompts that guide the model to focus on different reasoning objectives. This enables GPT to generate QA pairs that are faithful to the underlying scene context and adaptable across multiple evaluation criteria. The resulting dataset spans a range of reasoning types aligned with six predefined tasks introduced earlier.  After generating QA pairs, additional post-processing is conducted. It ensures the quality and format consistency of generated samples. This step emphasizes maintaining essential phrases for each task. For example, time grounding samples explicitly include phrases like "from frame A to frame B" to accurately localize events. This targeted filtering guarantees that each dataset element aligns precisely with the objectives of the defined spatio-temporal reasoning tasks.
An overview of the process is illustrated in Figure~\ref{fig:pipeline}.

\begin{table}[h]
\centering
\caption{Number of annotations per task and detailed statistics. We construct 178k annotations from 4D LiDAR, where each sequence yields 40 samples across six tasks, based on generated descriptions and human captions.}
\resizebox{0.8\columnwidth}{!}{
    \begin{tabular}{l|c|c|c}
        \toprule
        \textbf{Statistics} & \textbf{Train} & \textbf{Test} & \textbf{Total} \\
        \midrule
        Scene Description & 4,200 & 900 & 5,100 \\
        Max Sequence Length & 19 & 19 & 19 \\
        Min Sequence Length & 5 & 5 & 5 \\
        \midrule
        Existence & 18,545 & 3,770 & 22,315 \\
        Yes/No Q\&A & 37,026 & 7,525 & 44,551 \\
        Time Grounding & 13,124 & 2,783 & 15,907 \\
        Description & 18,540 & 3,770 & 22,310 \\
        Temporal Understanding & 23,956 & 4,757 & 28,713 \\
        Comprehensive Reasoning & 37,080 & 7,540 & 44,620 \\

        \midrule
        \textbf{Total annotation count} & \textbf{148,271} & \textbf{30,145} & \textbf{178,416} \\
        \bottomrule
    \end{tabular}}
    
    \label{tab:dataset_statistics}
    \Description[]{We designed the generation process to maximize dataset size while ensuring task diversity and minimizing redundancy.}
\end{table}

\subsection{Dataset Statistics}
\label{sec:statistics}

The B4DL dataset, generated using our data construction pipeline, is built from 850 scenes sourced from nuScenes, with 700 scenes allocated for training and 150 for testing. This split captures a wide range of lighting conditions and spatial layouts, ensuring that diverse and challenging scenarios are evenly distributed across both subsets. Each scene is further segmented into six sequences containing 3 to 10 frames, sampled at a consistent interval of two keyframes, which corresponds to one second given the 2Hz LiDAR frame rate. Consequently, the dataset comprises 4,200 sequences for training and 900 for testing, each paired with LiDAR point clouds, synchronized multi-view images, and numerically grounded manual annotations.

For each sequence, we generate 40 data samples to ensure a balanced distribution across all six reasoning tasks. Specifically, each sequence yields 5 samples for Existence, 10 for Binary QA, 5 for Time Grounding, 5 for Scene Description, 5 for Temporal Understanding, and 10 for Comprehensive Reasoning, which are subsequently refined through the post-processing step. This design promotes task diversity while maintaining consistency across sequences. As a result, the training set comprises approximately 14.8k samples, and the testing set includes 30.1k samples, each annotated with spatio-temporal QA pairs derived from the corresponding 4D LiDAR sequence. Detailed statistics are summarized in Table~\ref{tab:dataset_statistics}.

\section{B4DL Model and Training Pipeline}

We propose the B4DL model, an MLLM specifically developed to align 4D LiDAR data with language models. This model enables effective utilization of our generated 4D LiDAR dataset, the B4DL dataset, to understand the spatio-temporal dynamics present in real-world scenes. Notably, it supports LiDAR-only training by leveraging 3D point clouds and metadata captured solely by the LiDAR sensor. The model consists of three key modules—\textit{Encoder for LiDAR Point Cloud}, \textit{LiDAR Aligner}, and \textit{Metatoken}—which together allow the language model to process 4D LiDAR data in its high-dimensional form. The carefully designed training pipeline for the B4DL model adopts a two-stage strategy that progressively guides the model to comprehend and reason over 4D LiDAR scenes in a structured manner, enhancing its ability to perform spatio-temporal understanding tasks in complex outdoor environments. Our architecture builds on the pretrained large language model and integrates textual and 4D LiDAR modalities using cross-entropy loss for causal language modeling, facilitating joint training over both modalities. An overview is illustrated in Figure~\ref{fig:architecture}.

\subsection{4D LiDAR Alignment Modules}
\label{sec:architecture}

\noindent \textbf{Encoder for LiDAR Point Cloud}, denoted as \(E_L\), transforms 3D LiDAR point clouds into embeddings aligned with textual representations via CLIP \cite{radford2021learning}. While conventional 3D encoders are typically optimized for object detection, \(E_L\) is designed to project LiDAR data into a shared vision-language space by leveraging CLIP’s cross-modal alignment. Inspired by LiDARCLIP \cite{hess2024lidarclip}, we employ an encoder that leverages voxelization and sensor calibration to ensure consistent spatial alignment across frames. The resulting embeddings are used to compute the following similarity loss between LiDAR and image features:

\begin{equation}
\mathcal{L}_{\text{similarity}} = \frac{1}{d} (z_I - z_L)^T (z_I - z_L),
\end{equation}

\noindent where \( z_I, z_L \in \mathbb{R}^d \). Unlike the contrastive loss in CLIP training, this formulation uses only positive pairs, allowing us to leverage CLIP’s prior knowledge without the overhead of contrastive sampling. The encoder directly processes 3D point clouds:

\begin{equation}
    S_L = \{ P_1, P_2, \dots, P_T \}, \quad P_t \in \mathbb{R}^{N_t \times 4},
\end{equation}

\noindent where \( S_L \) denotes a sequence of LiDAR frames over time, and each frame \( P_t \) contains \( N_t \) points with spatial coordinates \((x, y, z)\) and intensity. Each frame \( P_i \) is encoded as:

\begin{equation}
    \{ p_i^{cls}, p_i^1, \dots, p_i^k \} = E_L(P_i), \quad i = 1, 2, \dots, T,
\end{equation}

\noindent where \( p_i^{cls} \) captures global spatial information, and \( \{ p_i^1, \dots, p_i^k \} \) represent localized features. The global embeddings \( p_i^{cls} \) across all frames are concatenated to form the 4D LiDAR Embedding, which serves as the input modality for the B4DL model.

\noindent \textbf{LiDAR Aligner}, \(f_p\), is implemented as a single linear layer that refines encoder-generated embeddings for seamless integration with LLMs. Although \(E_L\) produces CLIP-aligned embeddings, these may lack the granularity required for detailed language modeling, as CLIP primarily emphasizes high-level semantic alignment \cite{radford2021learning}. The aligner \(f_p\) maintains the high-dimensional structure of the input while projecting it into a space compatible with language-based reasoning, thereby functioning as a key interface between vision-language features and the LLM input space.

\noindent \textbf{Metatoken} is introduced to incorporate sensor metadata into the model. Denoted as `<meta>', it precedes a textual description of the ego vehicle equipped with the LiDAR sensor. Since LLMs struggle with raw decimal values due to tokenizer limitations, we convert sensor data into relative direction, position, velocity, and acceleration with respect to previous frames. This enables the model to capture temporal dynamics and causality. To avoid redundancy across sequences, we include textual descriptions only for the initial and final frames, which are concatenated after the `<meta>' token. Please refer to the appendix for the detailed formulation.

\subsection{Training Pipeline for B4DL Model}
\label{sec:training pipeline}

\begin{figure}[t]
    \centering
    \includegraphics[width=0.45\textwidth]{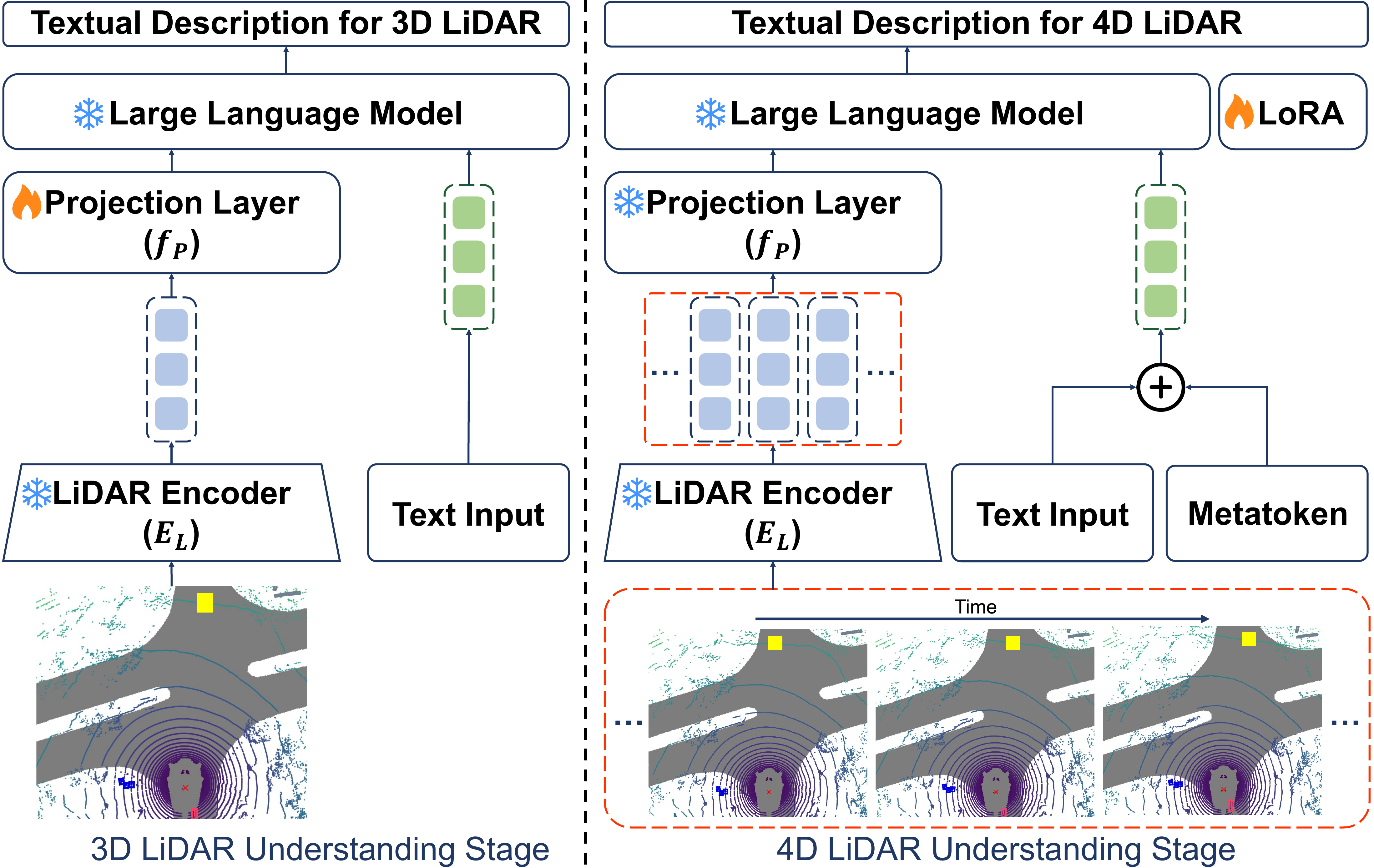}
    \caption{Visualization of the training pipeline. The projection layer \(f_p\) and a LoRA module are trained in two stages respectively, while other components remain frozen. The 4D LiDAR input is marked with a red dotted box.}
    \Description[]{We divide the training stage into two stages, and utilize Meta Token in the second stage.}
    \label{fig:architecture}
\end{figure}

To enable our B4DL model to perceive 4D LiDAR and perform spatio-temporal reasoning for 4D tasks, we structure the training process into two stages: \textit{3D LiDAR Understanding Stage} and \textit{4D LiDAR Understanding Stage}. Each stage progressively refines the model’s ability to first capture spatial relationships in 3D point clouds and then reason about spatio-temporal dynamics in 4D LiDAR. This staged approach ensures the model builds a strong spatial foundation before integrating language-based reasoning. A detailed illustration is provided in Figure~\ref{fig:architecture}.

\noindent \textbf{3D LiDAR Understanding Stage} focuses on training the LiDAR aligner \(f_p\) to enable the LLM to interpret LiDAR data by projecting the point cloud embedding \(p_i^{\text{cls}}\) into the LLM’s text embedding space. Although \( f_p \) is trained using only static 3D point clouds, the projection layer enables it to process them even in a concatenated form. As such, this projection layer is also trained and optimized to handle 4D LiDAR embeddings for the next stage.

Since this stage does not require temporal information, training is conducted using the LiDAR-LLM dataset, which consists of textual descriptions paired with static 3D point clouds from nuScenes. During this phase, only \(f_p\) remain trainable, while all other modules, including the encoder \(E_L\), are kept frozen. Note that the LiDAR-LLM dataset is paired with the point cloud embeddings to focus on the spatial information contained in the LiDAR data.

\noindent \textbf{4D LiDAR Understanding Stage} enhances the model’s ability to associate events with temporal progression, which is uniquely captured in 4D data but not in static 3D. To focus on temporal reasoning, we introduce a LoRA module \cite{hu2022lora}, while freezing all other components, including the encoder \(E_L\) and the LiDAR aligner \(f_p\). During training, the token `<4DLiDAR>' is prepended to all QA inputs to explicitly indicate that the task involves reasoning over 4D LiDAR data, guiding the model to attend to inherent spatio-temporal relationships within the input.

At this stage, the model receives the 4D LiDAR Embedding alongside aligned QA pairs from the B4DL dataset, constructed to maintain spatio-temporal consistency with the LiDAR sequences. To support temporal understanding, we introduce the term \textit{`frame'} to denote specific time steps and incorporate \textit{Metatoken}, previously described in Section~\ref{sec:architecture}, to provide context on the ego vehicle’s motion. The model is trained on both \textit{Simple Tasks} and \textit{Complex Tasks} as introduced in Section~\ref{ssec:task} where the \textit{Simple Tasks} guide learning of temporal changes and object dynamics, and the \textit{Complex Tasks} enhance the model’s capacity for higher-level reasoning grounded in dynamic scenes. Together, these tasks enable the MLLM to develop a structured, context-aware, and human-interpretable understanding of 4D LiDAR environments.

\begin{table*}[ht]
    \centering
    \caption{Quantitative performance comparison across different MLLM models on simple and complex Tasks. The best metric scores are highlighted. \dag denotes evaluation on the Waymo dataset to assess cross-dataset generalization of the B4DL model trained solely on our B4DL training set constructed from nuScenes.}
    \resizebox{.8\linewidth}{!}{
    \begin{tabular}{l|cc|ccccc}
        \toprule
        \multirow{2}{*}{\textbf{Model}} & 
        \multicolumn{2}{c|}{\textbf{Simple Tasks}} & 
        \multicolumn{5}{c}{\textbf{Complex Tasks}} \\
        \cmidrule(lr){2-3} \cmidrule(lr){4-8}
        & Accuracy $\uparrow$ & mIoU $\uparrow$ & B@4 $\uparrow$ & ROUGE-L $\uparrow$ & METEOR $\uparrow$  & BERTScore $\uparrow$ & GPT Score $\uparrow$\\
        \midrule
        B4DL-LiDARLLM & 0.611&	-&	0.018&0.189&	0.187&	0.868 & 42.425\\
        VTimeLLM \cite{huang2024vtimellm} & 0.694 & 0.160 & 0.083 &	0.305 & 0.262 & 0.893 & 55.654\\
        B4DL(Ours\textsuperscript{\dag}) & {0.706}&	{0.294}&	{0.053}	&{0.269}	&{0.228}	&{0.886} & 58.248\\
        
        \rowcolor{gray!20}B4DL(Ours) & \textbf{0.762}&	\textbf{0.311}&	\textbf{0.095}	&\textbf{0.322}	&\textbf{0.275}	&\textbf{0.897} & \textbf{59.513}\\
        
        \midrule
        
    \end{tabular}
}
    
    \Description[]{The table presents a quantitative comparison with other models, where our model outperforms all others across every metric}
    \label{tab:results}
\end{table*}

\section{Experiments}

\subsection{Experimental Settings}

We adopt Vicuna-7b-v1.5 \cite{zheng2023judging} as a large language model, which remains frozen during both training and inference. It serves as the backbone LLM throughout the entire pipeline. To facilitate the staged training pipeline of the B4DL model, we pre-train the encoder \( E_L \) using nuScenes, which provides 3D point clouds paired with multi-view images. For the image encoder, we adopt the CLIP model, specifically the ViT-L/14 variant \cite{radford2021learning}.

We use separate datasets for each training stage. For the \textit{3D LiDAR Understanding Stage}, we use the LiDAR-LLM-Nu-Caption dataset \cite{yang2023lidar}, which contains 162k QA pairs based on static LiDAR frames, to capture spatial relationships in point clouds. For the \textit{4D LiDAR Understanding Stage}, we use our B4DL dataset introduced in Section~\ref{sec:dataset}, consisting of 178k QA pairs designed for spatio-temporal reasoning over dynamic LiDAR sequences. Both datasets are derived from nuScenes and follow the same data split described in Section~\ref{sec:statistics}, with 700 scenes for training and 150 for testing. This consistent split prevents test data leakage. All experiments are run on a single NVIDIA RTX 4090 GPU and complete within 24 hours.

For evaluation, each metric is computed per task and then averaged to obtain the final score. Specifically, the accuracy of \textit{Simple Tasks} is the average of Existence and Binary QA scores. mIoU is based on the Time Grounding task, while the remaining four metrics come from \textit{Complex Tasks}.

\begin{figure*}[t]
    \centering
    \includegraphics[width=0.80\textwidth]{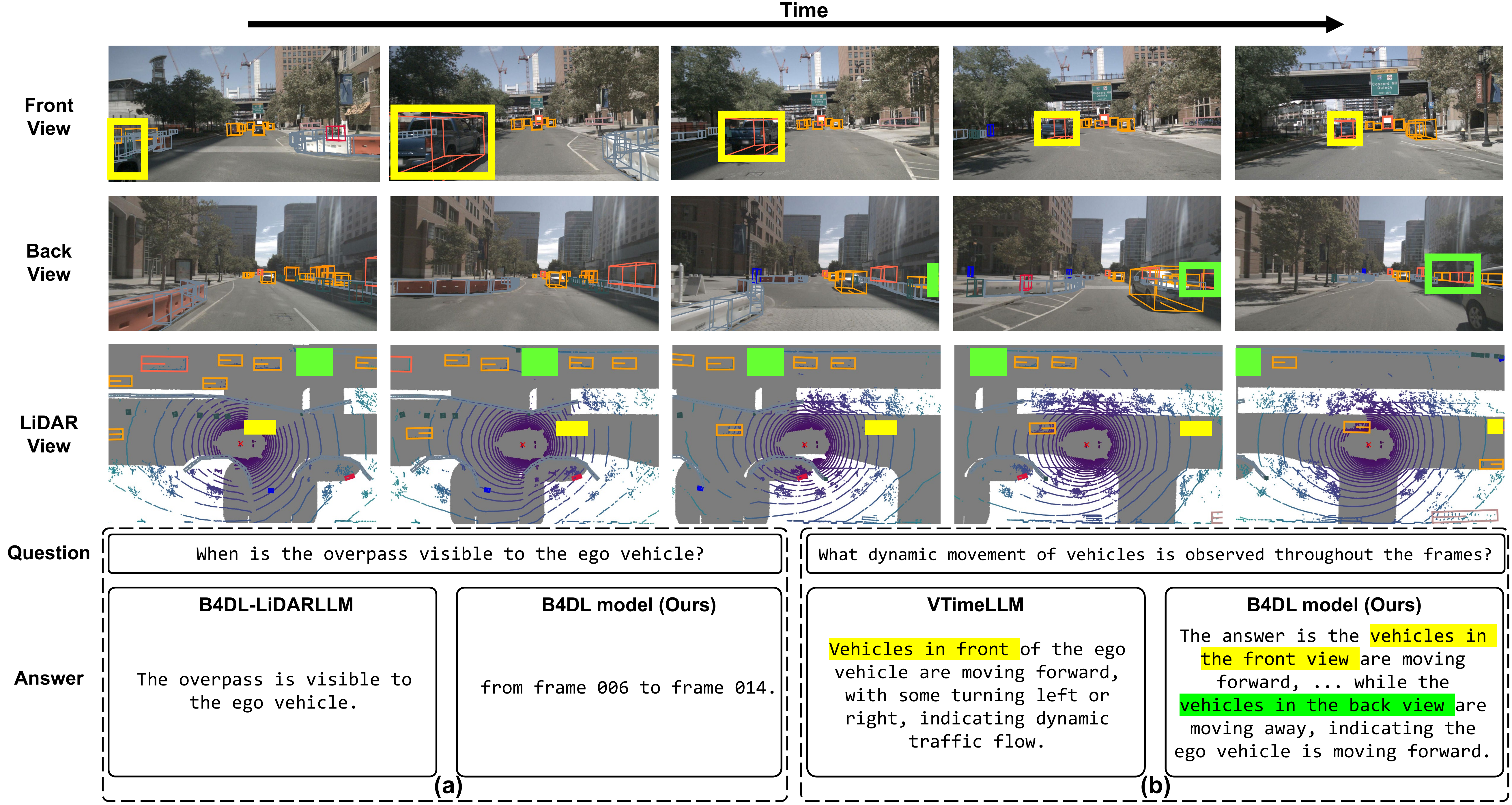}
    \caption{Qualitative comparison of generated answers from different MLLMs, including B4DL-LiDARLLM (a) and VTimeLLM (b). For qualitative perception, the LiDAR image, along with front and back view RGB images, are visualized. Yellow and green highlight the locations of the vehicle of interest, from the front and back perspectives respectively. (a) illustrates enhanced temporal understanding while (b) emphasizes spatial comprehension, showcasing how B4DL model outperforms others.}
    \Description[]{We conducted a comparison between our model and other existing models, demonstrating its spatio-temporal understanding capability.}
    \label{fig:qualitative}
\end{figure*}

\subsection{Main Results}
To demonstrate the unique capability of B4DL model to jointly understand 3D spatial structures and temporal dynamics, we conduct two comparative evaluations. First, to highlight the temporal advantages of 4D LiDAR over traditional 3D LiDAR, we compare our model with B4DL-LiDARLLM, a variant of the B4DL model trained only on the single-frame LiDAR-LLM dataset \cite{yang2023lidar}.
Second, to assess the spatial reasoning capability, we compare our model with VTimeLLM, a video-based MLLM capable of temporal reasoning.

Table~\ref{tab:results} presents the quantitative results comparing the models.
B4DL-LiDARLLM, which lacks temporal reasoning capability, does not generate time-related responses or frame intervals in a way that satisfies the requirements for mIoU evaluation. Furthermore,
as most of the test dataset demands both spatial and temporal understanding, its performance drops notably on the \textit{Complex Tasks}. Especially, B4DL-LiDARLLM showed a considerable performance drop in the GPT-based scoring, which evaluates semantic alignment and linguistic variation. This suggests that its failure in temporal understanding led to semantic mismatches in time-sensitive responses. Compared to VTimeLLM, our model achieves consistently better results across all tasks. In particular, it shows a significant improvement in mIoU, which can be attributed to its omnidirectional coverage. While VTimeLLM is limited to front or rear views, the B4DL model incorporates 360° LiDAR data to reason about the entire scene. Additionally, leveraging sensor metadata through \textit{Metatoken} provides contextual signals about ego motion and object dynamics, which further enhances performance across tasks.

\begin{table*}[ht]
    \centering
    \caption{Quantitative ablation study on the \textit{B4DL model}, showing performance variations based on the presence of Human Annotation (HA) and Metatoken. Metrics for both simple and complex tasks are presented and highlighted for the best score.}
    \resizebox{.75\linewidth}{!}{
    \begin{tabular}{cc|cc|ccccc}
        \toprule
        \multirow{2}{*}{\textbf{HA}} &
        \multirow{2}{*}{\textbf{Metatoken}} &
        \multicolumn{2}{c|}{\textbf{Simple Tasks}} &
        \multicolumn{5}{c}{\textbf{Complex Tasks}} \\
        \cmidrule(lr){3-4} \cmidrule(lr){5-9}
        & & Accuracy $\uparrow$ & mIoU $\uparrow$ & B@4 $\uparrow$ & ROUGE-L $\uparrow$ & METEOR $\uparrow$ & BERTScore $\uparrow$ & GPT Score $\uparrow$\\
        \midrule
          &               & 0.756 & 0.218 & 0.067 & 0.291 & 0.237 & 0.890 & 56.904\\
        \checkmark &      & \textbf{0.763} & 0.161 & \textbf{0.095} & 0.321 & \textbf{0.275} & \textbf{0.897} & 58.741\\
          & \checkmark   & 0.757 & 0.310 & 0.067 & 0.289 & 0.230 & 0.889 & 56.918\\ 
        \rowcolor{gray!20}
        \checkmark & \checkmark & 0.762 & \textbf{0.311} & \textbf{0.095} & \textbf{0.322} & \textbf{0.275} & \textbf{0.897} & \textbf{59.513}\\ 
        \bottomrule
    \end{tabular}
    }
    \Description[]{This table shows the advantages of using human annotation in dataset generation and presents the results of an ablation study comparing the performance with and without the inclusion of Metatoken during training.}
    \label{tab:ablation}
\end{table*}

Figure~\ref{fig:qualitative} provides a qualitative comparison of B4DL-LiDARLLM, VTimeLLM, and our model. In Figure \ref{fig:qualitative}-(a), B4DL-LiDARLLM fails to answer questions requiring temporal grounding, instead returning a generic description of the scene. This suggests that while the existing dataset and model can perform spatial reasoning, it is not enough to acquire the ability to understand temporal information. In contrast, our model correctly identifies the relevant frame range, demonstrating its ability to reason over time. As shown in Figure \ref{fig:qualitative}-(b), both models capture temporal dynamics and respond appropriately to time-sensitive queries. However, VTimeLLM processes only the front view, limiting its ability to provide spatially comprehensive answers. In contrast, our model incorporates both front and back perspectives, enabling reasoning over the full omnidirectional spatial context around the ego vehicle. This broader view allows our model to extract richer and more nuanced information.

\subsection{Ablation Study}

In this section, we analyze the impact of the key components introduced in B4DL, including both the B4DL dataset and the B4DL model. We also address critical concerns and provide corresponding clarifications in the following discussion.

\noindent \textbf{Generalizability to a Different Dataset} is evaluated by applying our data generation pipeline to the Waymo Open Dataset \cite{sun2020scalability} using the same methodology as nuScenes. Unlike nuScenes, which provides front and rear camera views, Waymo offers five front-facing cameras (front, front-left, front-right, side-left, side-right), enabling 180° forward coverage through lateral alignment. To accommodate this, we replaced the front and rear views used in the nuScenes setup with left and right views during data generation.

We evaluated our B4DL model, trained solely on the nuScenes-based B4DL dataset, by running inference on a 1k-sample subset of Waymo without any re-training or adaptation. As shown in Table~\ref{tab:results}, our model outperforms B4DL-LiDARLLM and VTimeLLM on the \textit{Simple Tasks}, demonstrating strong generalization capability. A moderate performance drop was observed on the \textit{Complex Tasks}, likely due to the prompt variation and scene framing differences between the lateral-view layout of Waymo and the front-rear configuration of nuScenes. Nonetheless, these results demonstrate the generalizability of our pipeline and its potential for seamless application to other LiDAR datasets. While some datasets may benefit from minor view-specific adjustments to accommodate differences in sensor configuration, our method offers a robust and adaptable foundation for deployment across diverse real-world settings.

\noindent \textbf{Necessity of Human Annotation (HA) and Metatoken} is thoroughly evaluated in Table~\ref{tab:ablation}, which reports how performance varies depending on the presence of HA during dataset synthesis and \textit{Metatoken} during the \textit{4D LiDAR Understanding Stage}. The best overall performance is achieved when both are included. HA improves the quality of synthesized QA pairs with manual supervision, while \textit{Metatoken} provides contextual cues on the ego vehicle’s motion, enhancing the model’s ability to learn spatio-temporal dynamics.

Removing HA leads to a substantial performance drop, especially in the \textit{Complex Tasks}, with notable degradation in B@4, METEOR, and ROUGE-L scores. These results indicate that HA is essential for producing responses that align lexically and semantically with human annotations, guiding the model to generate more structured and natural outputs. In contrast, removing \textit{Metatoken} has a smaller effect on overall metrics but causes a significant drop in mIoU, which reflects performance in the Time Grounding task. This highlights the role of \textit{Metatoken} in temporal localization by providing motion-related metadata that helps the model disambiguate dynamic interactions within the scene. Without it, the model struggles to distinguish whether changes in object distance result from ego motion, object motion, or both.

When both \textit{HA} and \textit{Metatoken} are removed, the model shows the lowest or near-lowest scores across most metrics, as expected. However, the mIoU is marginally higher than in the setting where only HA is included without \textit{Metatoken}. This suggests that, in the absence of motion context, \textit{HA} alone may introduce ambiguity by emphasizing static object information. In this case, the model may rely more directly on raw LiDAR embeddings, resulting in slightly better temporal localization than when partial supervision is applied without supporting temporal cues. These findings collectively confirm that both \textit{HA} and \textit{Metatoken} play complementary and essential roles in enabling robust spatio-temporal understanding.

\begin{table}[h]
    \centering
    \caption{DCScore \cite{zhu2025dcscore} comparison across different datasets. The best score is highlighted.}
    \resizebox{.5\linewidth}{!}{
    \begin{tabular}{l c}
        \toprule
        \textbf{Dataset} & \textbf{DCScore $\uparrow$} \\
        \midrule
        LiDAR-LLM     & $1.514 \pm 0.006$ \\
        NuScenes-QA   & \textbf{$1.982 \pm 0.006$} \\
        \rowcolor{gray!20}
        B4DL (train)  & $1.946 \pm 0.008$ \\
        \rowcolor{gray!20}
        B4DL (test)   & $1.948 \pm 0.008$ \\
        \bottomrule
    \end{tabular}
    }
    \label{tab:dcscore}
\end{table}

\noindent \textbf{Diversity of the Generated Dataset} may raise concern, as large-scale synthetic datasets can prioritize quantity over variety. To address this, we evaluate the diversity of the generated B4DL dataset using DCScore \cite{zhu2025dcscore}, a metric designed to assess the diversity of LLM-generated datasets. For fairness and reproducibility, we randomly sampled 1k QA pairs from each dataset and repeated the evaluation 100 times due to memory constraints. As shown in Table~\ref{tab:dcscore}, both the training and test splits of the textual B4DL dataset demonstrate strong diversity, achieving scores comparable to or even exceeding those of other datasets derived from the same source, nuScenes. Notably, compared to LiDAR-LLM, the B4DL dataset achieves significantly higher DCScore, likely due to the increased variation introduced by temporal reasoning. These results confirm that our generation pipeline not only scales effectively but also produces diverse and temporally grounded data, making it well-suited for real-world multimodal reasoning tasks.

\section{Conclusion}

In this work, we introduce B4DL, a comprehensive benchmark for understanding the spatio-temporal dynamics inherent in the real world by leveraging MLLMs tailored for 4D LiDAR. We present a carefully designed tasks for the benchmark, aimed at evaluating the model's ability to understand the 4D context and perform reasoning based on a high-level understanding of spatio-temporal dynamics. Additionally, we propose a new dataset generation pipeline and introduce the B4DL dataset, designed to enable both training and evaluating MLLMs on 4D LiDAR. Alongside this, we present the B4DL model, the first MLLM for 4D LiDAR with a training pipeline. Together, we aim for B4DL to serve as a foundation for comprehending the spatio-temporal dynamics captured by 4D LiDAR in outdoor scenes, paving the way for future advancements in understanding real-world dynamics leveraging 4D LiDAR.

\clearpage
\begin{acks}
This work was partly supported by the Institute of Information \& Communications Technology Planning \& Evaluation(IITP) grant funded by the Korea government(MSIT) (No.RS-2024-00439020, Developing Sustainable, Real-Time Generative AI for Multimodal Interaction, SW Starlab) and partly supported by the Institute of Information \& Communications Technology Planning \& Evaluation(IITP) grant funded by the Korea government(MSIT) (No.RS-2025-02283048, Developing the Next-Generation General AI with Reliability, Ethics, and Adaptability)
\end{acks}

\bibliographystyle{ACM-Reference-Format}
\bibliography{sample-base}

\clearpage

\appendix

\section{Project Page}
We provide a project page that offers rendered videos of 4D LiDAR data along with their associated textual descriptions generated by our proposed method, and inference results of our models. To comply with the anonymity policy, we have created a fully anonymized project page. The page last updated at 2025.04.11 23:54 PM AoE, before the submission deadline. The latest version of the project page is also included as a supplementary material attachment. You can check our project page at the following URL: \url{https://github.com/ccho4702/B4DL}

\section{Extended Related Works}
\subsection{3D Vision-Language Models}
With the recent advancements in multimodal language models, there has been increasing interest in extending their capabilities beyond 2D data to include 3D representations, including point clouds, neural fields, and meshes. Prior to the development of 3D LLMs, studies \cite{hegde2023clip, zhang2023clip, mohammad2022clip} explored aligning 3D data with text in a shared representation space, following CLIP’s approach. PointCLIP \cite{zhang2022pointclip} achieved 3D recognition by projecting point clouds into multi-view depth maps, while PointCLIP V2 \cite{zhu2023pointclipv2} enhanced performance with GPT-4 \cite{achiam2023gpt} generated 3D-specific text, extending its capabilities to segmentation and detection. LidarCLIP \cite{hess2024lidarclip} leveraged image-LiDAR pairs to train a point cloud encoder, using the image domain as an intermediary to align LiDAR and text representations.

Following the success of these methods, various 3D data-driven LLMs have emerged \cite{zhu20233dvista, wang2023beyond, huang2023chatscene, tang2024minigpt}. Among them, 3D-LLM \cite{hong20233d} extracts 2D features from multi-view images rendered from 3D scenes to construct 3D representations, enabling context-aware responses to language prompts. It supports diverse 3D assets and tasks, including captioning, question answering, grounding, and dialogue. LiDAR-LLM \cite{yang2023lidar} builds a 3D LiDAR-text paired dataset to fine-tune a view-aware transformer and adapter, facilitating captioning, grounding, and question answering with LiDAR data. However, these 3D MLLMs focus primarily on static 3D scenes and lack the capability to comprehend dynamic real-world environments.

\subsection{Language Models for Sequential Data}
Research on the temporal understanding of LLMs for sequential data has primarily focused on video LLMs \cite{li2023videochat, maaz2023videochatgpt, yuan2019findtalk, wang2023internvid}. VTimeLLM \cite{huang2024vtimellm} is designed for fine-grained temporal understanding, employing boundary-aware training to improve event boundary detection in videos. VideoLLaMA \cite{damonlpsg2023videollama} is an instruction-tuned multimodal model that integrates visual and auditory information using a vision-language and audio-language branch. By combining pre-trained encoders with a query-based transformer, it aligns multi-modal features with LLMs. Through pre-training and instruction tuning, it generates context-aware responses grounded in both modalities. While these models demonstrate temporal understanding in sequential data, they struggle to capture spatial interactions within 3D space.

\subsection{Video-Based Datasets}

The integration of spatial and temporal reasoning has become crucial in real-world applications, particularly in domains like autonomous driving. Therefore, the importance of datasets for enabling spatio-temporal understanding of a given scene has become increasingly emphasized. Several previous studies, such as DriveGPT-4 \cite{xu2024drivegpt4}, DriveLM \cite{sima2023drivelm}, and LingoQA \cite{marcu2024lingoqa}, have introduced datasets and benchmarks for visual question answering (VQA) based on multi-frame video inputs in autonomous driving. While these datasets, composed of video and text, are useful for training LLMs to understand temporal dynamics, they present challenges in enabling LLMs to capture the spatial features of 4D real-world environments.

\subsection{LiDAR-Based Datasets}
LiDAR-LLM \cite{yang2023lidar} introduced a 3D LiDAR-text pairing dataset and an MLLM for 3D scene understanding. While its dataset enables spatial reasoning from LiDAR-based point clouds, its single-frame nature limits temporal comprehension. NuScenes-QA \cite{qian2023nuscenes} incorporates multi-frame LiDAR data but relies on short sequences and template-generated answers, reducing variability and limiting robust instruction tuning on MLLMs. It also lacks annotations for spatio-temporal changes, such as object motion and directional shifts, which are essential for comprehensive 3D scene reasoning. In contrast, B4DL dataset is designed as a LiDAR-specific, multi-frame sequential dataset with annotations for 360-degree coverage and temporal sequences, making it well-suited for MLLM training.


\section{Detailed Metatoken Formulation}

\begin{figure*}[t]
    \centering
    \includegraphics[width=0.92\textwidth]{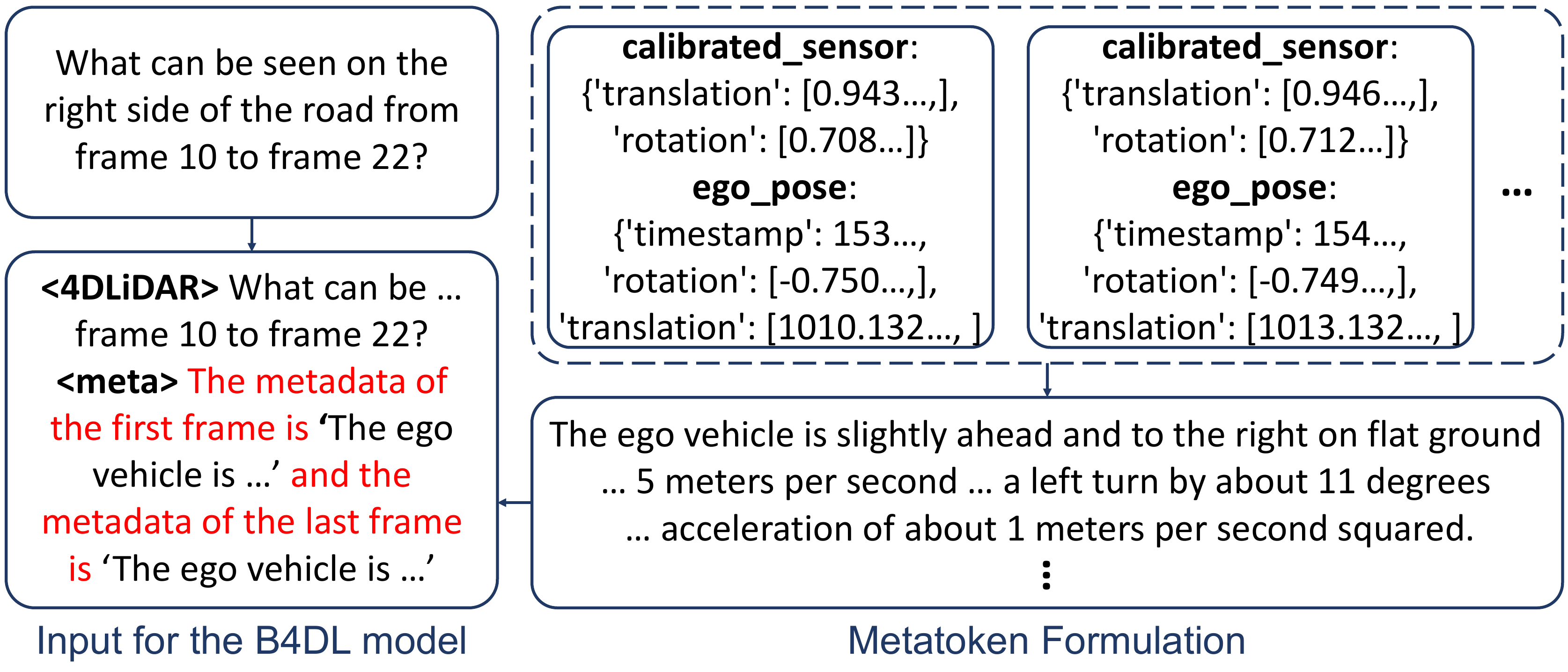}
    \caption{Visualization of metatoken construction and a sample B4DL input. Raw numbers are converted to text, with the first and last frame texts joined by red-highlighted words to form the metatoken.}
    \Description[]{We divide the training stage into two stages, and utilize Meta Token in the second stage.}
    \label{fig:metatoken}
\end{figure*}

As illustrated in Figure~\ref{fig:metatoken}, each LiDAR point cloud is associated with raw metadata representing the calibrated sensor’s positional state, the ego vehicle’s motion, and a timestamp indicating time progression. Since these values follow a consistent numerical format, we convert them into textual descriptions that capture velocity, direction, position, and acceleration relative to previous frames. Once the \textit{Metatokens} are formulated, they are used as part of the input to the B4DL model. Specifically, the input question is prepended with <4DLiDAR>, followed by <meta>, after which the metatoken descriptions of the first and last frames referenced in the QA pair are concatenated. These are joined using the red-highlighted connective phrases shown in Figure~\ref{fig:metatoken}. The resulting text sequence is then fed into the B4DL model.

\section{Additional Comparison to MLLM of other modalities}
To evaluate our model's capability in spatio-temporal reasoning and 4D scene understanding, a benchmark or dataset specifically designed for 4D dynamics would ideally facilitate meaningful comparisons. However, currently, no publicly available benchmark or dataset with accessible code directly addresses our specific needs. Even existing datasets involving sequences of 3D data typically focus only at the object level, and thus they are inadequate for comprehensive evaluation. Consequently, this limitation necessitates comparisons using alternative modalities.

\begin{table*}[ht]
    \centering
    \caption{Performance across different training scales on simple and complex tasks.}
    \label{tab:scale-performance}
    \resizebox{.7\linewidth}{!}{
    \begin{tabular}{l|cc|cccc}
        \toprule
        \multirow{2}{*}{\textbf{Scale (\%)}} & 
        \multicolumn{2}{c|}{\textbf{Simple Tasks}} & 
        \multicolumn{4}{c}{\textbf{Complex Tasks}} \\
        \cmidrule(lr){2-3} \cmidrule(lr){4-7}
        & Accuracy $\uparrow$ & mIoU $\uparrow$ & B@4 $\uparrow$ & METEOR $\uparrow$ & ROUGE-L $\uparrow$  & BERTScore $\uparrow$\\
        \midrule
        10 & 0.741 & 0.168 & 0.082 & 0.306 & 0.258 & 0.893  \\
        25 & 0.750 & 0.298 & 0.089 & 0.313 & 0.266 & 0.895 \\
        50 & 0.757 & 0.310 & 0.094 & 0.320 & 0.273 & 0.896  \\
        75 & 0.761 & \textbf{0.312} & 0.094 & 0.320 & 0.272 & 0.896  \\
        100 & \textbf{0.762} & 0.311 & \textbf{0.095} & \textbf{0.322} & \textbf{0.275} & \textbf{0.897}  \\
        \bottomrule
    \end{tabular}
    }
\end{table*}

\section{Scaling Effects and Data Sufficiency}
Table~\ref{tab:scale-performance} presents the model’s performance across varying training scales, from 10\% to 100\%. As the amount of training data increases, performance steadily improves across both simple and complex tasks, with gains gradually saturating near full-scale training. This trend not only highlights the effectiveness of our model in leveraging more data, but also demonstrates that the overall dataset is sufficiently large to support robust and generalizable learning, particularly for handling complex reasoning tasks.

\section{Prompt Design for Dataset Generation and QA Creation}

To explain the prompt engineering behind our B4DL dataset construction, we present examples of the prompts used in the dataset generation process.
Table~\ref{tab:generate_description} shows the full prompt used to generate descriptions from multi-view frontal images. It is designed to focus on 4D LiDAR data, capturing temporal changes in the 3D point cloud rather than static appearance.
Table~\ref{tab:qa-instruction} presents the prompt used for generating question-answer pairs from the generated descriptions. This prompt takes into account both front and back scene descriptions and instructs GPT to produce QA pairs that reflect spatio-temporal understanding, including frame grounding and directional context. Table~\ref{tab:gen_example} contains an example of generated descriptions. It also includes human annotations, start index, and end index.

\section{Additional Qualitative Results}

To further highlight the quality of the generated \textit{B4DL dataset}, we present several qualitative samples as references.



\begin{figure*}[!h]
    \centering
    \includegraphics[width=1\textwidth]{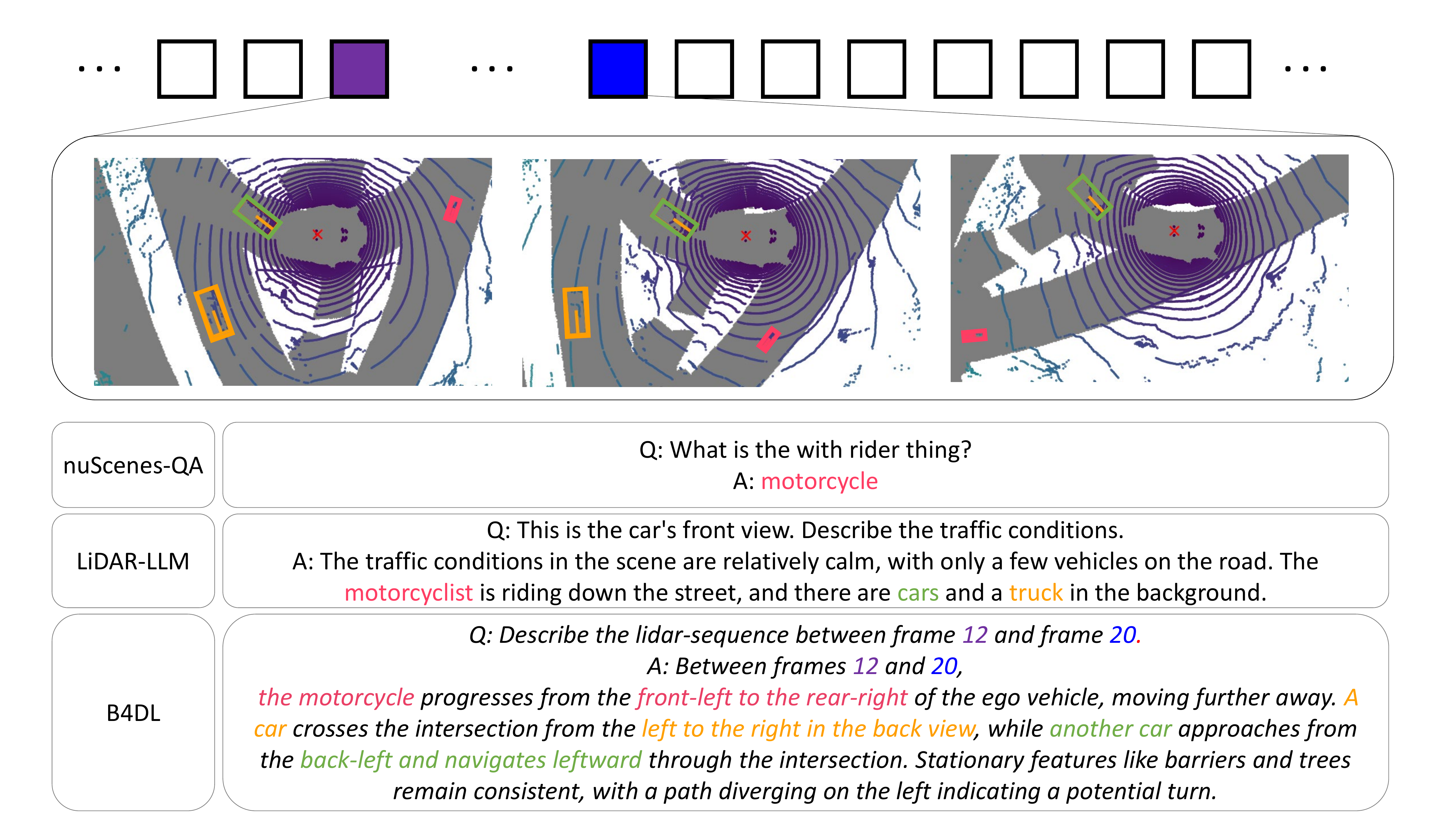}
    \caption{Comparison of textual datasets for LiDAR data in nuScenes dataset.}
    \Description[]{Fully described in the text.}
    \label{fig:dataset_comparison}
\end{figure*}

In Figure~\ref{fig:dataset_comparison}, we compare our B4DL dataset with two other datasets based on the LiDAR point cloud data from nuScenes. Our B4DL dataset outperforms the others in spatial detail and exhibits superior temporal understanding. While the nuScenes-QA and LiDAR-LLM datasets focus primarily on the presence of objects in static point clouds, the B4DL dataset captures not only the presence of objects but also their changes over time, reflecting 4D spatio-temporal dynamics.

\begin{figure*}[!h]
    \centering
    \includegraphics[width=0.9\textwidth]{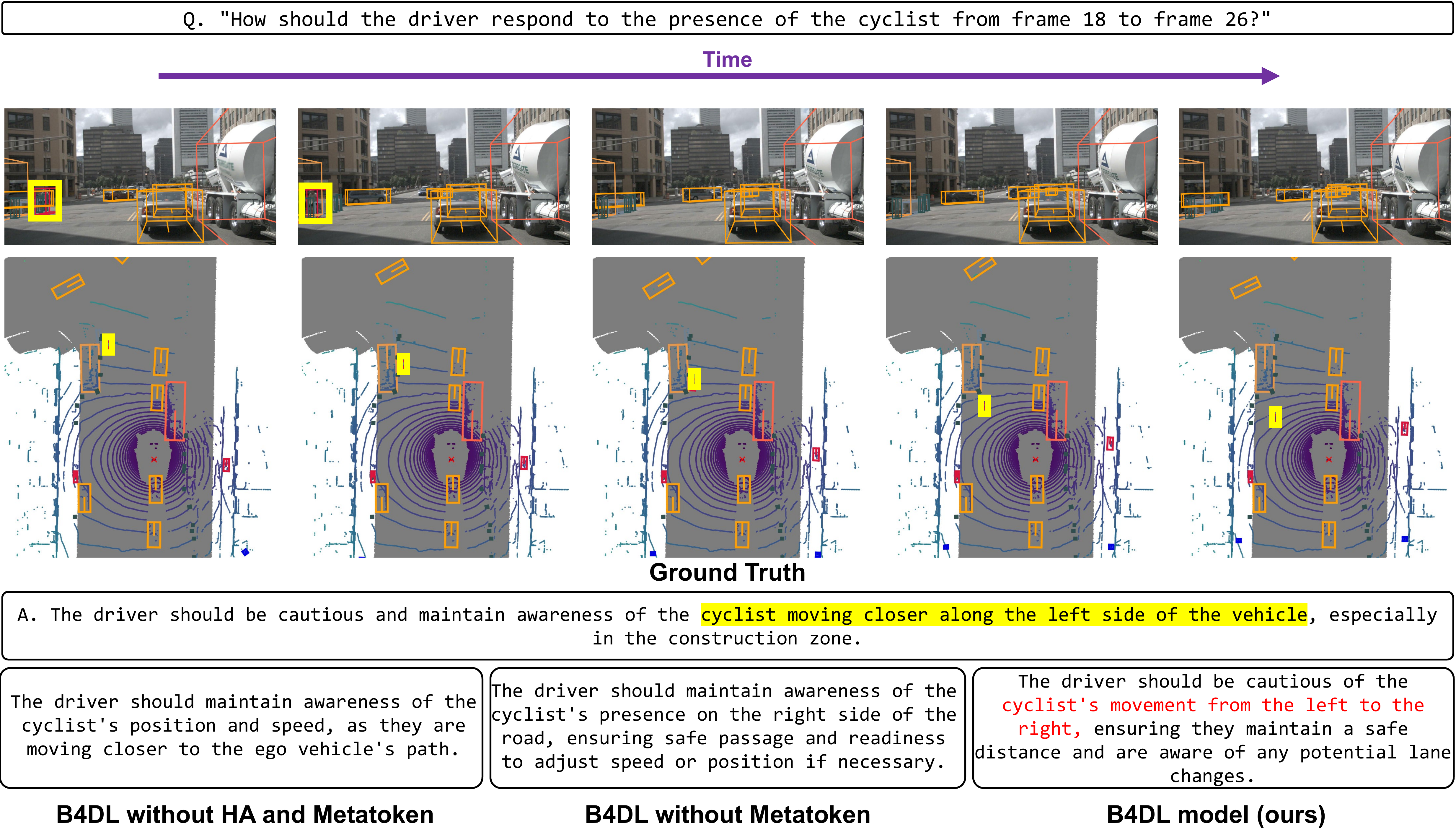}
    \caption{Ablation comparison within B4DL model for Human Annotations (HA) and Metatoken.}
    \Description[]{Fully described in the text.}
    \label{fig:ablation qualitative 1}
\end{figure*}

Figure~\ref{fig:ablation qualitative 1} presents a qualitative analysis from the ablation studies conducted on our B4DL model. We observed that, in the absence of HA and Metatoken, the B4DL model is only able to capture the broad spatial changes in the scene. This is evident as the object highlighted in yellow is consistently traced across LiDAR images and identified in all three cases shown in the figure. However, the model struggles with directionality and distinguishing whether the observed spatial change in the object corresponds to actual movement. In other words, without HA and Metatoken, the model is prone to confusing relative motion, as it lacks the full context of the ego vehicle's status. This highlights the significance of the proposed modules, HA and Metatoken, which provide critical guidance to the B4DL model, resulting in more accurate and sophisticated responses.

\begin{figure*}[!h]
    \centering
    \includegraphics[width=1\textwidth]{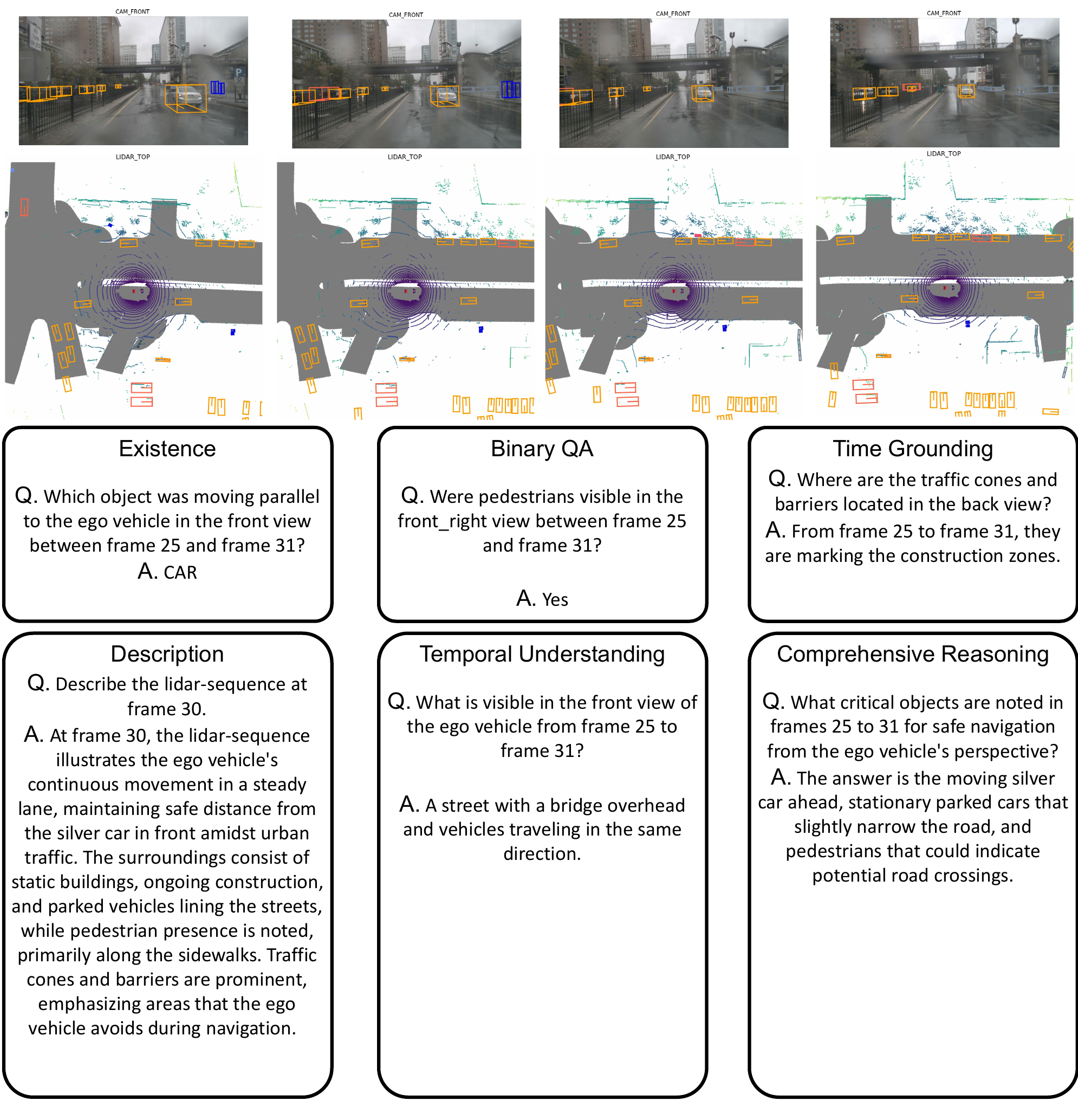}
    \caption{Extra examples of the generated dataset for 6 different tasks.}
    \Description[]{Fully described in the text.}
    \label{fig:tasks}
\end{figure*}

\begin{figure*}[!h]
    \centering
    \includegraphics[width=1\textwidth]{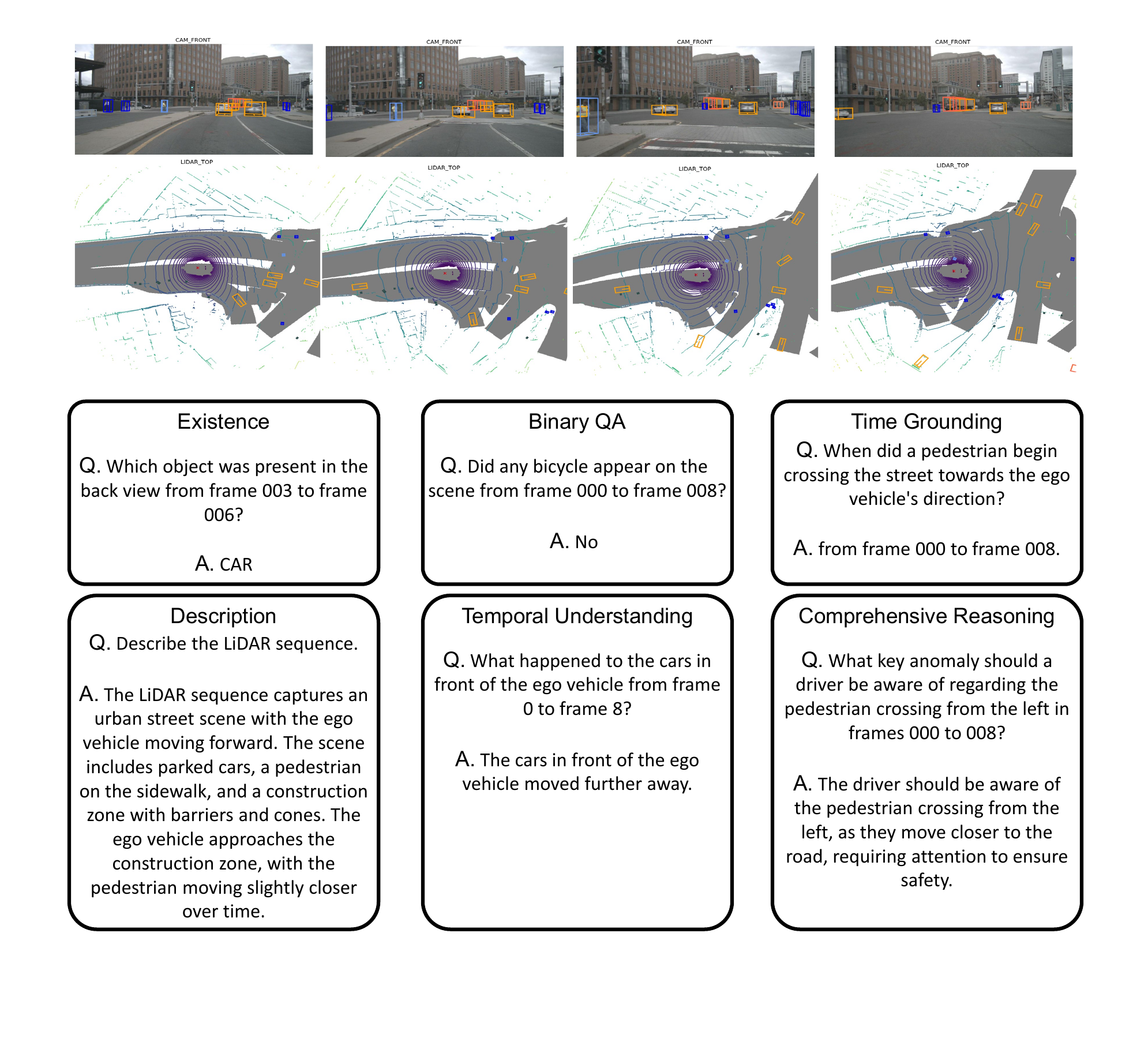}
    \caption{Extra inference results for 6 difference tasks.}
    \Description[]{Fully described in the text.}
    \label{fig:inference_output}
\end{figure*}

In Figure~\ref{fig:tasks}, we display six generated sample QA datasets from the tasks we have meticulously designed. These descriptions account for omnidirectional changes and identify temporal variations. These examples are randomly selected from the our synthesized B4DL dataset. Finally, in Figure~\ref{fig:inference_output}, we present the actual inference outputs in the form of QA pairs. We examined all six tasks and showcase the answers obtained through inference. These answers align well with the observed events in the scene, as seen in the corresponding camera and LiDAR images.

\clearpage

\begin{table*}[t]
\centering
\caption{Instruction used for generating 4D LiDAR point cloud descriptions from multi-view image sequences.}
\label{tab:generate_description}
\begin{tabular}{>{\RaggedRight\arraybackslash}p{2.5cm}|>{\RaggedRight\arraybackslash}p{14.5cm}}
\hline
\textbf{Instruction} &
You are a helpful assistant that generates captions for sequences of frames to analyze the corresponding 3D point cloud captured by LiDAR.

\par\medskip
\textbf{INSTRUCTIONS:}\par
-- The first \{frame\_len\} frames are from the front view, the next \{frame\_len\} from the front\_left, and the last \{frame\_len\} from the front\_right.\par
-- Generate a single-paragraph description of the scene captured in the LiDAR point cloud by integrating information from all three views (front, front\_left, front\_right).\par
-- Focus on object types, relative positions, shapes, sizes, distances, and movements. Do not include color, text, lighting, weather, or other 2D-specific details.\par
-- Emphasize temporal changes, mentioning frame numbers, directions (left, right, front, back), and whether objects are approaching or moving away.\par
-- Structure the response into three parts:\par
\quad [1] Description of the Scene\par
\quad [2] Key Changes Over Time\par
\quad [3] Important Objects and Events from the Driver's Perspective\par
-- LiDAR can only classify objects into the following categories: \{Animal, pedestrian, stroller, wheelchair, barrier, debris, trafficcone, construction, motorcycle, bicycle, car, bus, trailer, truck, suv\}. Do not infer color, text, or semantic content.\par
-- Mention any special movements of the ego vehicle, if applicable.\par
-- The frames are as follows: \texttt{\{list of frames here\}}\\
\hline
\end{tabular}
\end{table*}

\begin{table*}[t]
\centering
\caption{Instruction used for generating question and answer pairs from front and back multi-frame descriptions.}
\label{tab:qa-instruction}
\begin{tabular}{>{\RaggedRight\arraybackslash}p{2.5cm}|>{\RaggedRight\arraybackslash}p{14.5cm}}
\hline
\textbf{Instruction} &
You are a helpful assistant that creates question and answer pairs using the description of the front and back views of multi-frame scenes.

\par\medskip
\textbf{INSTRUCTIONS:}\par
-- This is a description of the front parts of the ego vehicle from frame \{start\_index\} to frame \{end\_index\}: \{front\_description\}.\par
-- This is a description of the back parts of the ego vehicle from frame \{start\_index\} to frame \{end\_index\}: \{back\_description\}.\par
-- Ensure the model considers the back view correctly: objects on the left correspond to the ego vehicle’s right, and objects on the right correspond to its left, based on the LiDAR perspective.\par
-- Generate 10 Q\&A pairs that comprehensively describe the scene, considering spatial relationships, object interactions, and possible dynamics.\par
-- Include temporal information such as the corresponding frame number using the format “from frame 000 to frame 000”.\par
-- Each Q\&A pair should follow this format:\par
\quad Q: What is the question? \par
\quad A: The answer is the answer.\par
-- Only provide the Q\&A pairs without any other embellishments.\par
-- Do not use numbering such as Q1, A1, etc.\par
-- The following is part of the ground truth for this sequence: \{gt\_description\}. Use this information if necessary.\\
\hline
\end{tabular}
\end{table*}

\begin{table*}[t]
    \centering
    \caption{GPT scoring instructions used for semantic evaluation.}
    \label{tab:placeholder}
    \begin{tabular}{>{\RaggedRight\arraybackslash}p{2.5cm} | >{\RaggedRight\arraybackslash}p{14.5cm}}
        \hline
        \textbf{Instruction} &
        You are an expert evaluator for semantic answer quality. You will be given a set of question-answer-ground truth (Q/A/GT) triplets. Your task is to evaluate how semantically close the given answer (A) is to the ground truth (GT), and how well it responds to the question (Q).\par
        Focus only on:\par
        -- Whether the answer conveys the same or similar meaning as the ground truth.\par
        -- Whether the answer correctly and sufficiently addresses the question.\par
        -- Do not consider wording, phrasing, or grammar unless they change the meaning.\par
        Score the answer from 0 to 100:\par
        -- 100: Fully aligned with GT in meaning, complete and relevant.\par
        -- 80–99: Mostly aligned with minor omissions or slight deviation.\par
        -- 60–79: Partially aligned; captures key ideas but misses or misrepresents some.\par
        -- 40–59: Limited relevance or meaning overlap with GT.\par
        -- 20–39: Barely related to the GT.\par
        -- 0–19: Meaningless or unrelated.\par
        \vspace{1ex}
        \textbf{EXAMPLES:}\par
        Question: What changes occur to the car in front over the frames?\par
        Answer: The car in front moves slightly forward from frame 000 to frame 008.\par
        GT: The car in front moves slightly forward until it stops again from frame 0 to frame 8.\par
        Score: 90\par
        \vspace{0.5ex}
        Question: From frame 004 to frame 006, how does the building structure in the back view affect the scene?\par
        Answer: The building structure on the left side of the road in the back view remains constant, providing a stable reference point for the ego vehicle's navigation.\par
        GT: The building structure remains fixed, providing a static backdrop in the scene without impacting the dynamics.\par
        Score: 100\par
        \vspace{0.5ex}
        Question: How do the cars to the right change over time from frame 00 to frame 16?\par
        Answer: The cars to the right remain stationary, indicating they are parked.\par
        GT: They become slightly closer to the ego vehicle.\par
        Score: 50\par
        \vspace{0.5ex}
        Question: What moving vehicles are visible in the ego vehicle's path by frame 026?\par
        Answer: A car is moving away from the ego vehicle.\par
        GT: There are no significant moving vehicles in the visible path.\par
        Score: 10\par
        \vspace{1ex}
        \textbf{INSTRUCTIONS}\par
        Question: \{q\} \quad GT: \{ref\} \quad Answer: \{pred\}\par
        Please provide a score between 0 and 100 based on the quality of the prediction compared to the reference. The score should reflect how well the prediction aligns with the reference in terms of semantic similarity and relevance to the question. Only provide the score without any additional text. \\
        \hline
    \end{tabular}
\end{table*}

\begin{table*}[t]
\centering
\caption{An example of generated description and human annotation.}
\label{tab:gen_example}
\begin{tabular}{p{2.5cm}|p{14.5cm}}
\hline
\textbf{Example} &

\textbf{Scene Token:} 01c8c59260db4a3682d7b4f8da65425e \par
\textbf{Sequence ID:} 101198145 \par

\textbf{Description (Front View):} \par
[1] \textbf{Description of the Scene:} The 3D point cloud captures an urban street scene. In the front view, the road extends forward with buildings on both sides. On the left side of the road, there are pedestrians near the entrance of a building, and a bus parked near the loading bay slightly further up. The street has several lamp posts and trees lining the left sidewalk. In the frontal right view, there are larger buildings across the street with clearly marked pedestrian pathways. \par

[2] \textbf{Key Changes Over Time:} Over the frames, the pedestrians on the left move slightly, some coming closer to the entrance, indicating minor lateral movement. The bus remains stationary near its parking position. As frames progress from front to front\_left and to front\_right views, the distance to the built structures remains mostly constant suggesting the ego vehicle is maintaining its position. However, the relative positioning indicates slight lateral movement to monitor the environment. \par

[3] \textbf{Important Objects and Events from the Driver's Perspective:} The primary concern from the driver's perspective includes the group of pedestrians near the building on the left, as any sudden movement onto the road could be critical. The parked bus on the left side poses as a fixed object. Continuous monitoring of the sidewalk and the road for potential crossing events by pedestrians is crucial. Objects such as trees and streetlamps form part of environmental navigation cues and maintaining orientation. \par

\textbf{Description (Back View):} \par
[1] \textbf{Description of the Scene:} The LiDAR point cloud captures a scene featuring a multi-lane roadway flanked by greenery and structures. Overhead, a large pedestrian overpass spans the road, with support pillars visible on either side. On the road and surrounding areas, various objects such as barriers, traffic cones, and vehicles like cars and buses are present, spaced at varying distances from the ego vehicle, which maintains a centered position on the road. To the sides, there are shrubs and structural elements like stairways and buildings. \par

[2] \textbf{Key Changes Over Time:} From frames 12 to 20, the overhead pedestrian bridge becomes increasingly prominent, indicating that the ego vehicle is approaching and passing directly beneath it. Vehicles in the scene, especially buses and cars, are stationary but slightly change position relative to the ego vehicle, suggesting minor forward movement. Barriers and traffic cones maintain consistent alignment with road markings, showing minor lateral movement as the vehicle aligns itself post-bridge. \par

[3] \textbf{Important Objects and Events from the Driver's Perspective:} Key objects include the pedestrian overpass directly overhead, multiple stationary cars and a bus on the right, and structural barriers on both sides. These elements are crucial for navigation, as avoiding collisions is necessary. The presence of barriers and traffic cones requires careful maneuvering to stay within lane boundaries. The stationary bus, partly in a tunnel, suggests a point of interest or potential obstruction ahead that the driver must prepare for. \par

\textbf{Ground Truth Annotation:} \par
A human pedestrian police\_officer approached the ego vehicle between Frame 015 and Frame 016. \par

\textbf{Start Index:} 12\par
\textbf{End Index:} 20
\\
\hline
\end{tabular}
\end{table*}

\end{document}